\PassOptionsToPackage{table,dvipsnames}{xcolor}
\documentclass[10pt,twocolumn,letterpaper]{article}
\usepackage{cvpr}              % To produce the CAMERA-READY version
\usepackage{graphicx}
\usepackage{amsmath}
\usepackage{amssymb}
\usepackage{multicol}
\usepackage{float}
\usepackage{mathptmx} % 使用 Times New Roman 字体
\usepackage{booktabs}
\usepackage{xspace}

\newcommand{\alias}{G3Flow\xspace}

\definecolor{bestcolor}{gray}{.9}
\newcommand{\bestcell}[1]{\cellcolor{bestcolor}{#1}}

\newcommand{\myparagraph}[1]{\vspace{3pt}\noindent\textbf{#1}}

\usepackage{float}

\usepackage{changepage}

\usepackage{amsmath}
\usepackage{amssymb}
\usepackage{mathtools}
\usepackage{pifont}
\usepackage{amsthm}
\theoremstyle{plain}

\theoremstyle{definition}

\theoremstyle{remark}

\usepackage{algorithm}
\usepackage{algpseudocode}
\algrenewcommand\algorithmicrequire{\textbf{Input:}}
\algrenewcommand\algorithmicensure{\textbf{Output:}}

\usepackage{listings} % 用于代码列表
\usepackage{fontawesome} % 用于特殊符号

\newcommand{\R}{\mathbb{R}}

\definecolor{hku_color}{HTML}{13A983} % 注意HTML颜色代码是大写的
\definecolor{szu_color}{HTML}{84193E} % 同上
\definecolor{table_color}{HTML}{CCCCFF}
\definecolor{darkblue}{rgb}{0, 0.12, 0.55}
\definecolor{darkgreen}{rgb}{0, 0.55, 0.12}
\definecolor{darkred}{rgb}{0.6,0,0}
\definecolor{darkgreen}{rgb}{0,0.6,0}
\definecolor{Gray}{gray}{0.9}
\definecolor{Acolor}{RGB}{126,171,85}   % A的颜色
\definecolor{Bcolor}{RGB}{79,113,190}   % B的颜色
\definecolor{Ccolor}{RGB}{104,52,154}   % C的颜色
\definecolor{Dcolor}{RGB}{164,109,62}   % C的颜色

\usepackage[breaklinks=true,
            colorlinks,
            linkcolor = darkred,
            urlcolor  = magenta, 
            citecolor = teal,
            bookmarks = false]{hyperref}
\newcommand{\ours}[0]{\alias}
\title{\textbf{\textcolor{szu_color}{G}3\textcolor{hku_color}{F}low: \textcolor{szu_color}{{G}}enerative 3D Semantic \textcolor{hku_color}{{F}}low for Pose-aware and \\Generalizable Object Manipulation}}
\author{
Tianxing Chen$^{1,2,3}$\footnotemark[1] \quad
Yao Mu$^{1}$\footnotemark[1]\hspace{1pt} \footnotemark[2] \quad 
Zhixuan Liang$^1$\footnotemark[1] \quad 
Zanxin Chen$^{3,4}$ \quad 
Shijia Peng$^{3,4}$\quad
Qiangyu Chen$^{3}$\\
Mingkun Xu$^{5}$\quad 
Ruizhen Hu$^{3}$\quad 
Hongyuan Zhang$^{1,2}$\quad 
Xuelong Li$^{2}$\quad 
Ping Luo$^{1,6}$\footnotemark[2]\\
[1.5mm]
$^1$The University of Hong Kong \quad
$^2$Institute of Artificial Intelligence (TeleAI), China Telecom\\
$^3$Shenzhen University \quad
$^4$AgileX Robotics \quad
$^5$GDIIST \\
$^6$HKU Shanghai Intelligent Computing Research Center
\\
\normalsize{\url{https://tianxingchen.github.io/G3Flow}}
}
\begin{document}

\maketitle

\footnotetext[1]{Co-first authors. Equal contribution.}
\footnotetext[2]{Corresponding authors. Contact {\tt \{ymu, pluo\}@cs.hku.hk}}

% teaser
% \twocolumn[{\renewcommand\twocolumn[1][]{#1}
% \maketitle
% \begin{center}
% \vspace{-21pt}
% \includegraphics[width=1.0\linewidth]{figures/teaser.jpg}
% \vspace{-18pt}
% \captionsetup{type=figure}
% \caption{\textbf{Teaser}}
% \end{center}
% }]

\begin{abstract}
Recent advances in imitation learning for 3D robotic manipulation have shown promising results with diffusion-based policies. However, achieving human-level dexterity requires seamless integration of geometric precision and semantic understanding. We present \textbf{\textcolor{szu_color}{G}3\textcolor{hku_color}{F}low}, a novel framework that constructs real-time semantic flow, a dynamic, object-centric 3D semantic representation by leveraging foundation models. Our approach uniquely combines 3D generative models for digital twin creation, vision foundation models for semantic feature extraction, and robust pose tracking for continuous semantic flow updates. This integration enables complete semantic understanding even under occlusions while eliminating manual annotation requirements. 
By incorporating semantic flow into diffusion policies, 
% we demonstrate significant improvements in both terminal-constrained manipulation and cross-object generalization.
%
% Recent advances in robotic manipulation have primarily focused on geometric representations, overlooking the crucial role of semantic understanding in achieving human-level manipulation capabilities. While diffusion-based policies have shown promising results in imitation learning, their reliance on purely geometric features limits their effectiveness in terminal-constrained tasks and intra-class generalization. We present \ours, a novel approach that leverages foundation models to generate and maintain dynamic 3D semantic fields for enhanced robotic manipulation. Our method uniquely combines three key components: (1) automated semantic field generation using 3D generative models and vision foundation models, eliminating the need for manual annotation; (2) occlusion-free object-centric semantic representation through digital twins integration; and (3) efficient real-time field maintenance using robust pose tracking with minimal camera requirements. By incorporating these semantic fields into diffusion policies, we demonstrate significant improvements in both terminal-constrained control tasks and intra-class generalization. 
%
extensive experiments across five simulation tasks show that \textbf{\textcolor{szu_color}{G}3\textcolor{hku_color}{F}low} consistently outperforms existing approaches, achieving up to 68.3\% and 50.1\% success rates on terminal-constrained manipulation and cross-object generalization respectively. Our results demonstrate the effectiveness of \textbf{\textcolor{szu_color}{G}3\textcolor{hku_color}{F}low} in enhancing real-time dynamic semantic feature understanding for robotic policies.

% \keywords{3D Semantic Fields, Robotic Manipulation, Foundation Models, Digital Twins, Diffusion Policy}
\end{abstract}  
\begin{figure}[h] 
    \centering
    \includegraphics[width=1.0\linewidth]{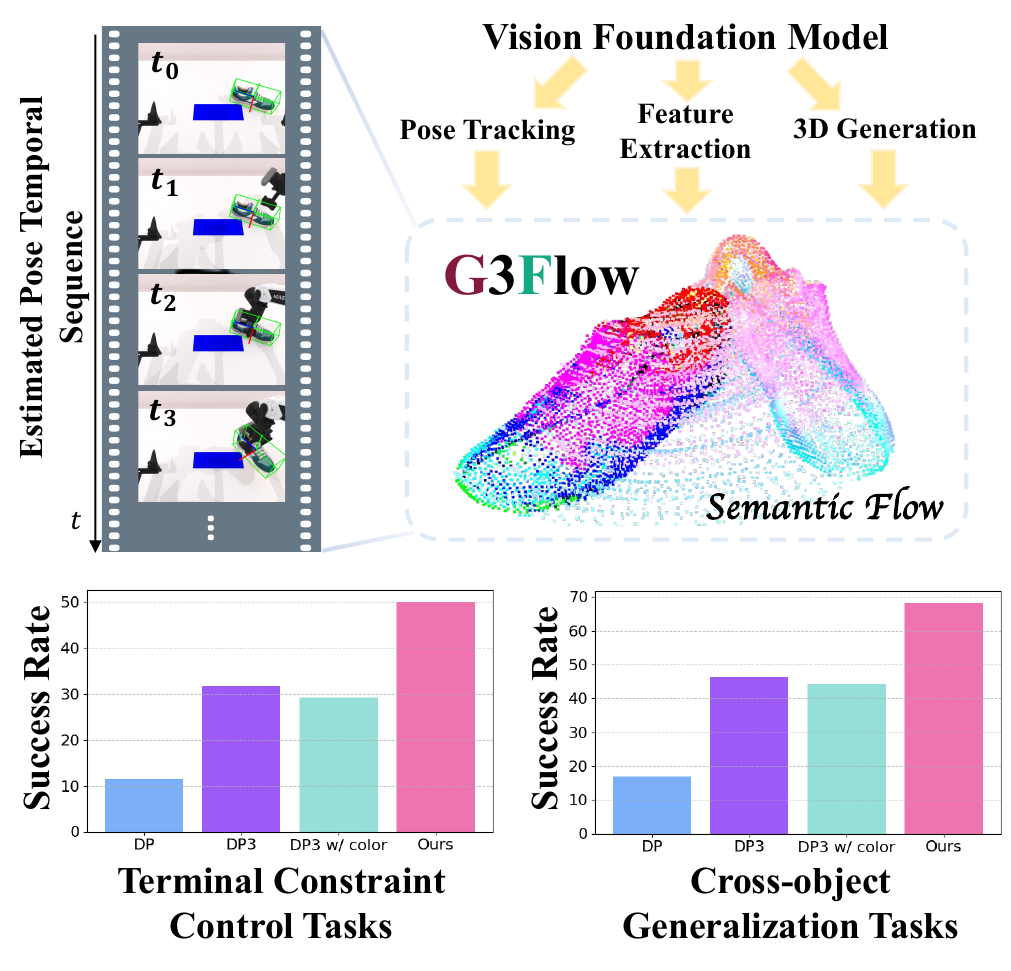}
    \vspace{-22pt}
    \caption{\textbf{Motivation of \ours.} Our approach leverages 3D generative model and language-guided detection model to generate 3D semantic flow (top). Through continuous field tracking, \ours enables pose-aware and generalizable manipulation, demonstrating superior performance across terminal constraint control and cross-object generalization tasks over multiple baselines (DP, DP3, and DP3 w/ color) (bottom).}
    \vspace{-15pt}
    \label{fig:teaser} 
\end{figure}
\section{Introduction}
\label{sec:intro}

% Recent years have witnessed remarkable progress in robotic manipulation. However, achieving advanced manipulation capabilities remains challenging, particularly in tasks requiring both precise control and adaptability - from accurately aligning tools with specific parts to handling variants of the same object category with consistent precision. These challenges stem from a fundamental limitation in current approaches: while they excel at capturing geometric properties, they lack comprehensive understanding of objects' functional parts and their semantic relationships, crucial for both precise manipulation and cross-object generalization. In the domain of imitation learning for robotics, diffusion-based policies have emerged as the de facto standard, demonstrating unprecedented performance in learning complex manipulation skills from demonstrations. However, following the broader trend in robotic manipulation, these policies primarily operate on geometric representations without explicit consideration of semantics.

Recent years have witnessed significant advances in imitation learning for robotic manipulation, leading to remarkable achievements across various tasks~\cite{zhao2023learning,chi2023diffusion,goyal2024rvt,lin2024data,liu2024rdt}. Image-based imitation learning methods often face challenges in precise manipulation and sample efficiency due to their limited ability to capture geometric relationships. In parallel, researchers have developed 3D imitation learning methods utilizing point clouds~\cite{ze20243d,ke20243d,gervet2023act3d} or voxel~\cite{grotz2024peract2,shridhar2023perceiver} representations to enhance few-shot learning capabilities by capturing geometric information. Among them, 3D diffusion policies~\cite{ze20243d,ke20243d} have shown promising results across multiple robotic tasks, owing to their superior ability to model multi-modal distributions. However, these geometry-centric methods, despite their advantages, often lack the crucial semantic understanding, necessary for sophisticated manipulation tasks. For instance, in pose-aware manipulation scenarios such as shoe placement and tool use, purely geometric representations struggle to identify semantically meaningful parts (such as the toe of a shoe) and hardly handle cases where semantically similar objects exhibit large geometric variations. This limitation highlights a critical research direction: the integration of rich semantic information from 2D images with geometric features from 3D representations. Such integration is essential for advancing performance in tasks that demand both precise spatial control and semantic comprehension, potentially bridging the gap between geometric precision and semantic understanding in robotic manipulation.

Several approaches have recently emerged to address this semantic understanding challenge in robotic manipulation. D$^3$Fields~\cite{wang2024d3fields} introduced dynamic 3D descriptor fields to enhance geometry representation, while subsequent works like GenDP~\cite{wang2024gendp} explored category-level generalization through semantic fields. However, these methods face significant practical challenges that they require manual keypoint selection and a multi-view setup for complete field generation and struggle with maintaining semantic consistency during dynamic interactions. 
%
% Specifically, occlusions during manipulation not only result in incomplete object observations but also pose significant challenges to feature acquisition, severely affecting semantic understanding and limiting their real-world applicability.

We propose \alias, a foundation model-driven framework that constructs real-time 3D semantic flow—an object-centric, occlusion-robust semantic representation using only a single-view camera without manual annotations. \alias combines 3D generative models for digital twin creation, foundation models for semantic feature extraction, and general pose trackers for continuous semantic updates. It operates in two phases: (1) active object-centric exploration to build a semantic field via 3D reconstruction, and (2) real-time semantic flow generation through pose tracking during manipulation. This representation enhances policy learning by providing consistent and complete semantic understanding.

% In this paper, we propose \alias, a novel foundation model-driven approach that constructs real-time 3D semantic flow, a dynamic, object-centric, and complete 3D semantic representation that maintains consistency even under occlusions. G3Flow is annotation-free and reconstructs complete object semantics using only a single-view camera with the help of foundation models. Our key insight is to leverage the 3D generative model to create precise digital twins from multiple observations, the detection foundation models to extract rich semantic features, and the general pose tracking models to enable continuous semantic flow updates. 
% % This combination eliminates manual annotation while ensuring persistent semantic understanding throughout the manipulation process.
% %
% Specifically, our framework operates in two phases: (1) Initial semantic field establishment through object-centric exploration and 3D object generation, where a robot actively gathers multiple observations to create a comprehensive digital twin and its base semantic field; and (2) Semantic flow generation through real-time pose tracking, continuously transforming the semantic field to create a dynamic flow that aligns with the physical object during manipulation, maintaining completeness even under occlusions. This semantic flow serves as a powerful representation for downstream policy, enabling it to better handle both precise control and object variations.

Through extensive experiments on both terminal-constrained manipulation and cross-object generalization tasks, we demonstrate that policy with \ours significantly outperforms existing methods, achieving up to 68.3\% v.s.~46.2\% and 50.1\% v.s.~31.7\% success rates comparing to next best method on these two kind tasks. Our approach achieves superior success rates in precise manipulation tasks and shows strong generalization capabilities across object variants, validating the effectiveness of our semantic flow framework. These results demonstrate the potential of enhancing imitation learning policies with rich semantic representations, paving the way for more precise and generalizable robotic manipulation.

Our key contributions can be summarized as follows: 
(1) We propose a novel foundation model-driven approach for constructing semantic flow, a dynamic and complete semantic representation through the integration of 3D generation, detection, and pose tracking models, enabling real-time understanding robust to occlusions without manual annotation.
(2) We develop a semantic flow-based imitation learning framework that leverages the dynamic semantic representation for enhanced manipulation, enabling both precise terminal control and effective generalization across object variations. 
% within similar categories. 
%
%
(3) Through extensive validations, we demonstrate our semantic flow significantly enhances imitation learning policies, achieving up to 68.3\% and 50.1\% success rates on terminal-constrained control and cross-object generalization tasks respectively.

% Code is available on https://github.com/TianxingChen/GSFields.
\section{Related Works}
\label{sec:related-work}

\subsection{3D Semantic Fields for Robotics Manipulation}
Semantic fields have emerged as a promising direction for enhancing robotic manipulation by providing rich semantic understanding of the environment~\cite{wang2024d3fields,qiu2024open,wang2024gendp,shen2023distilled}. These approaches bridge the gap between geometric perception and semantic comprehension.
% crucial for advanced manipulation capabilities.
%
Representatively, D$^3$Fields\cite{wang2024d3fields} firstly integrated dynamic 3D descriptor fields with manipulation. 
% and subsequent works further extended its potential. 
OVMM~\cite{qiu2024open} explored open-vocabulary mobile manipulation through vision-language models. And GenDP~\cite{wang2024gendp} addressed category-level generalization in diffusion policies. F3RM~\cite{shen2023distilled} enabled natural language specification through CLIP-based semantic distillation.

However, 
% fundamental challenges remain in obtaining and maintaining reliable semantic fields for robotic manipulation. 
current approaches like D$^3$Fields~\cite{wang2024d3fields} and GenDP~\cite{wang2024gendp} heavily rely on manual keypoint selection and have great difficulty in complete field generation when occlusion occurs while occlusion usually happens during manipulation due to the obstruction caused by robotic arms. It not only results in incomplete observation but also poses significant challenges to feature acquisition for downstream policy. 
Thus, \ours are proposed to address the limitations by a novel foundation model-driven approach, enabling dynamic, object-centric, and complete 3D semantic fields representation in real-time under closed-loop control.

\subsection{3D Generative Models for Robotic Simulation}
% Recent advances in 3D object generation have witnessed various foundational models employing different technical approaches. 
Recent advances have witnessed great progress in 3D object generation.
Early attempts like GET3D~\cite{gao2022get3d} leveraged generative adversarial networks to produce textured 3D meshes from images, while Point-E \cite{nichol2022point} and Shap-E \cite{jun2023shap} respectively explored text-to-3D generation through point clouds and implicit functions. Following these works, diffusion-based approaches such as DreamFusion \cite{poole2022dreamfusion} and Magic3D \cite{lin2023magic3d} demonstrated improved capability in synthesizing high-resolution 3D content from text descriptions.

However, these methods often struggle with generating intricate geometric details and high-fidelity textures, crucial for realistic robot manipulation. Rodin~\cite{xu2023rodin} addressed them with enhanced capabilities for producing detailed and textured 3D objects, validated by multiple practical robot tasks, such as RoboTwin~\cite{mu2024robotwin}, making it particularly suitable for our work in creating realistic virtual simulations.
% , where it facilitates the creation of digital twins for robotic training and simulation purposes.

\subsection{Diffusion Models for Imitation Learning}
Diffusion model~\cite{sohldickstein2015deep,DDPM} is a powerful class of generative models which models the score of data distribution (the gradient of energy function) rather than the energy itself~\cite{singh2023revisiting,salimans2021should}. The key idea behind diffusion models is their iterative transformation of a simple prior distribution into a target one through a sequential denoising process.
In robotics, diffusion-based policies \citep{chi2023diffusion, reuss2023goal,Ze2024DP3,pearce2023imitating,diffuser,diffusion_policy,metadiffuser,adaptdiffuser,skilldiffuser,lu2024manicm,liang2024dexdiffuser} have demonstrated impressive performance in learning complex manipulation skills from demonstrations. Recent works have explored various directions: 3D Diffusion Policy~\cite{Ze2024DP3} combines 3D scene representations with diffusion objectives, ChainedDiffuser~\cite{xian2023chaineddiffuser} focuses on trajectory generation between key-poses, and 3D Diffuser Actor~\cite{ke20243d} tackles joint key-poses and trajectory prediction.
However, these approaches primarily operate on geometric representations without explicit semantic understanding, limiting their precision in terminal-constrained manipulation and generalization across object variations. Our work \ours addresses this limitation by leveraging Foundation Models to maintain accurate and consistent semantic information during dynamic interactions, enabling more precise and generalizable manipulation capabilities.

\begin{figure*}[tb] 
    % \vspace{-10pt}
    \centering
    \includegraphics[width=0.95\textwidth]{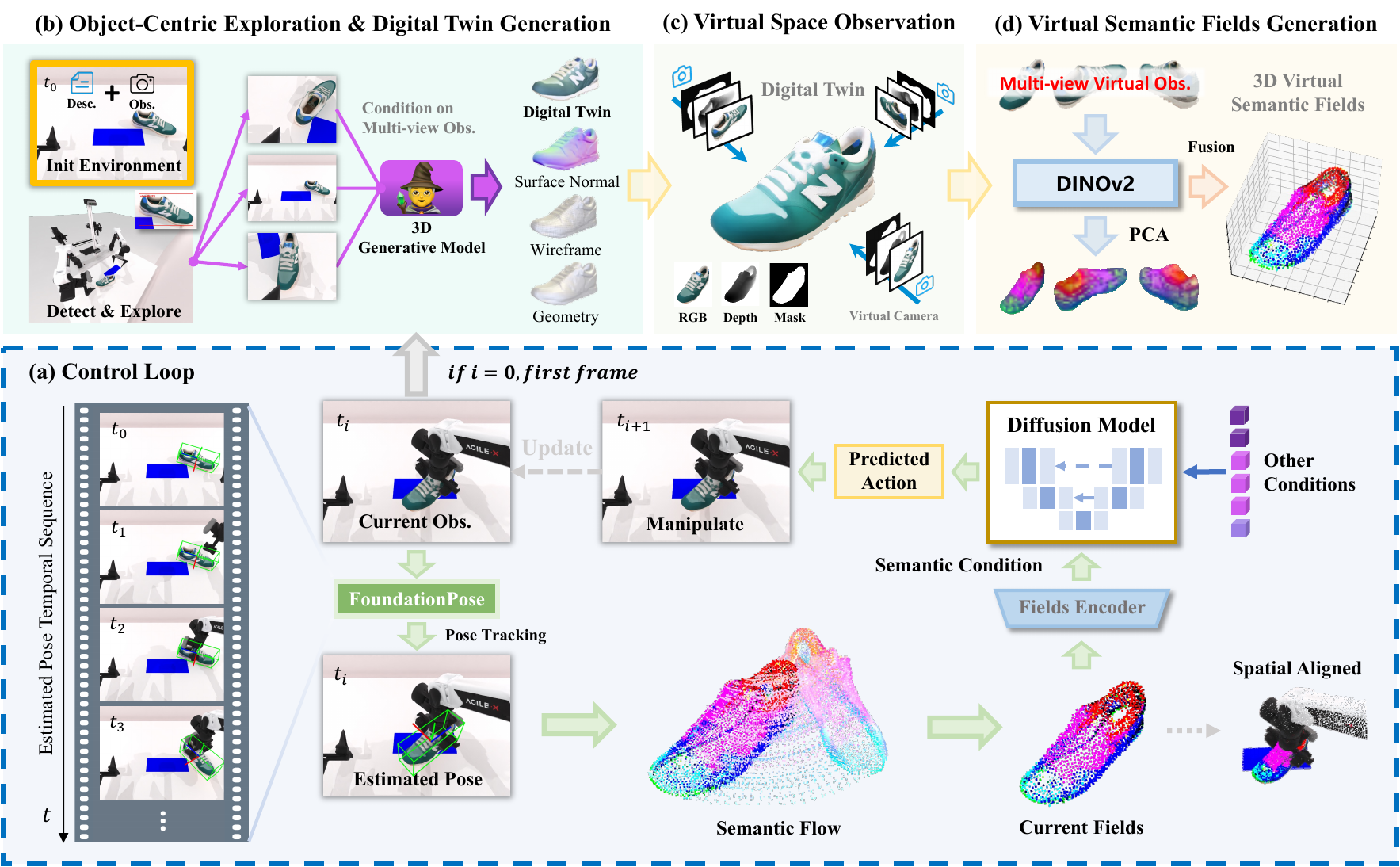}
    % \vspace{-3pt}

\caption{\textbf{Pipeline of \ours.} Our framework consists of (top) an initialization phase that generates comprehensive 3D representation (surface normals, wireframe, and geometry) through object-centric exploration and digital twin generation, which enables rich semantic field extraction, and (bottom) a control execution phase where real-time pose tracking maintains dynamic semantic fields to guide diffusion-based manipulation actions for pose-aware and generalizable manipulation.}

    \label{fig:main} 
    \vspace{-10pt}
\end{figure*}

\section{Method}
\label{sec:method}

% \caption{\textbf{Pipeline of \ours for pose-aware and generalizable object manipulation.} Our framework consists of initialization and control execution phases. In the initialization phase (top row), object-centric exploration gathers multi-view observations, which are processed by a digital twin generative model to create a comprehensive 3D representation with surface normals, wireframe, and geometry. This digital twin enables multi-view virtual observations, which are processed through DINOv2 and PCA to generate 3D virtual semantic fields. In the control execution phase (bottom row, dashed box), FoundationPose tracks object poses in real-time at each timestep $t_i$. The estimated pose transforms the semantic fields to maintain alignment with the physical object. These transformed fields serve as semantic conditions for a diffusion model to predict manipulation actions at $t_{i+1}$, ensuring consistent semantic understanding even under partial observations or occlusions.}

\subsection{Overview}
We formulate our problem as how to get and maintain semantic flow $\mathcal{O}{vsf}$, and how to learn a visuomotor policy $\pi:\mathcal{O}\mapsto\mathcal{A}$ from expert data, where the observation space $\mathcal{O}$ is composed of real point cloud observations $\mathcal{O}{r}$ and $\mathcal{O}{vsf}$. 
% and $\mathcal{A}$ denotes the action space of the robotic arm.
Our key insight is to leverage foundation models to construct and maintain complete 4D semantic understanding during dynamic interactions through real-time semantic flow, which addresses the limitations of existing geometry-centric approaches in handling occlusions and semantic variations.

Our framework operates in two phases: (1) Initial semantic flow construction through object-centric exploration and digital twin generation, where a robot actively gathers multi-view observations to create a comprehensive digital twin and extract its semantic features; and (2) Dynamic flow maintenance through real-time pose tracking, which continuously transforms these semantic features to align with physical objects during manipulation, maintaining completeness even under challenging occlusions or partial observations.
Specifically, we first employ a 3D generative model to reconstruct high-fidelity digital twins from multi-view RGB observations, leveraging the model's embedded knowledge to accurately infer even unseen object parts. The reconstructed twins enable semantic feature extraction through DINOv2~\cite{oquab2023dinov2} and dimensionality reduction via PCA~\cite{kurita2019principal} in a virtual environment, creating an initial semantic point cloud. We then utilize FoundationPose~\cite{wen2024foundationpose} for robust object pose tracking in real-world scenarios, enabling dynamic transformation of these semantic features while preserving completeness under occlusions and partial observations.

% Our approach leverages Foundation Models to create a novel 4D semantic flow framework. The pipeline begins with a 3D generative model that reconstructs a 3D asset of task-relevant  object from limited RGB observations, utilizing the model's embedded common sense to infer unseen object parts. By generating multi-view observations of this reconstruction and processing them through DINOv2~\cite{oquab2023dinov2}, we establish a comprehensive semantic field. The semantic features are refined through PCA~\cite{kurita2019principal} and projected into 3D space using virtual depth information, creating an initial semantic point cloud. The key innovation lies in how we transform this static representation into a dynamic 4D semantic flow: we employ FoundationPose~\cite{wen2024foundationpose} tracker to continuously track the object's pose in real-world scenarios, which allows us to dynamically transform the initial semantic point cloud. This approach maintains a complete and persistent semantic field even under challenging conditions like occlusions and complex manipulations.

Our system, \ours, consists of five key modules detailed in the following sections: a) Object-centric Exploration for active multi-view observation collection; b) Object 3D Model Generation through 3D generative models; c) Virtual Semantic Flow Generation combining digital twins with vision foundation models; d) Spatial Alignment via Object Tracking; and e) \ours-enhanced Diffusion Policy leveraging  both $\mathcal{O}{r}$ and $\mathcal{O}{vsf}$ for precise manipulation. Figure \ref{fig:main} illustrates our framework.

% Thus, our system, denoted as \ours, is divided into several modules: a) \textbf{Object-centric Exploration:} At the initial moment, the robotic arm is driven to actively move around the object to obtain multi-view observations; b) \textbf{Digital Twin Generation:} Using the multi-view observations obtained in (a), a precise digital twin of the object is created using a 3D Generative Model; c) \textbf{Virtual Semantic Fields Generation:} Combining the digital twin with a Vision Foundation Model (VFM) to obtain rich and complete object-centric virtual space semantic fields; d) \textbf{Spatial Alignment via Object Tracking:} Based on object tracking, the virtual space semantic fields can be spatially transformed to align with the object point cloud in real space; e) \textbf{\ours for Diffusion Policy:} Utilizing Diffusion Policy as the backbone, action sequences are generated based on $\mathcal{O}_{r}$ and $\mathcal{O}_{vsf}$ observations to drive the robotic arm to interact correctly with the environment. An overview of \ours is in Figure \ref{fig:main}. We will detail each part in the following sections.

\subsection{Initial Semantic Flow Construction}
\begin{figure}[h] 
    \centering
    \includegraphics[width=0.4\textwidth]{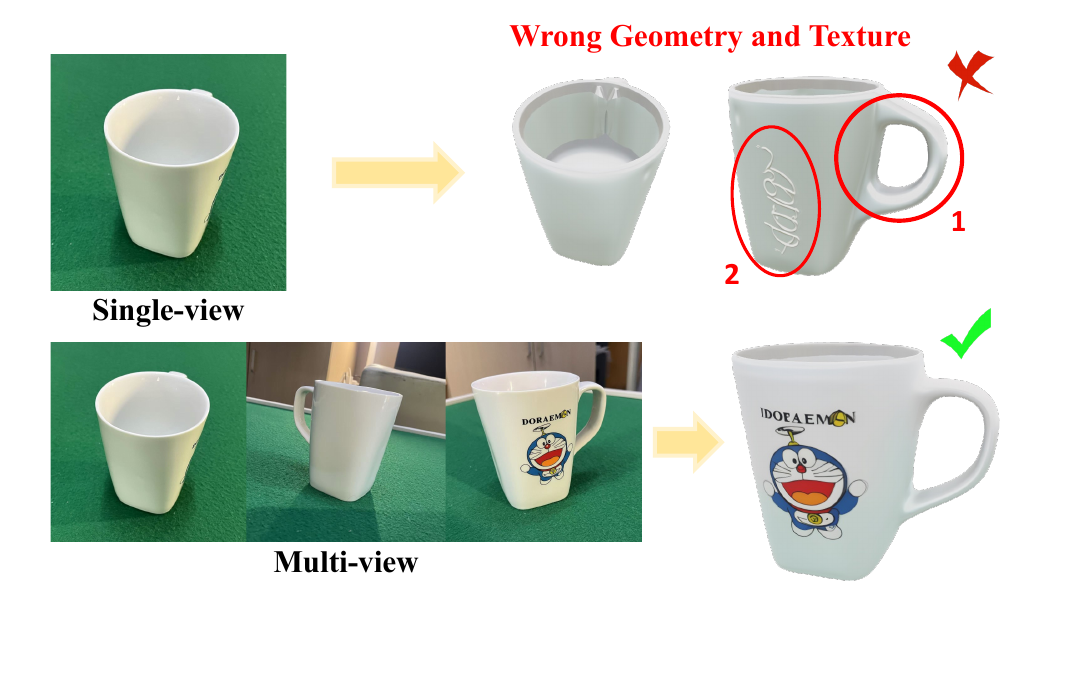}
    \vspace{-20pt}
    \caption{\textbf{Failure mode of single-view 3D generation.} When using a single view for 3D generation, certain geometric details may be inaccurately reconstructed due to occlusion, even if the result appears plausible from a commonsense perspective.}
    \vspace{-15pt}
    \label{fig:gen_teaser} 
\end{figure}

\myparagraph{Object-Centric Exploration.}
\label{sec:object-centric-exploration}
To construct an accurate and complete semantic flow, our first phase focuses on obtaining comprehensive object observations. Conventional single-view approaches face two critical challenges: First, poor initial object poses can lead to incomplete reconstructions due to self-occlusions (\eg, a mug's handle being hidden from the camera view). Second, during manipulation, the robot arm often occludes the camera's view of the target object, resulting in information loss. As shown in Fig.~\ref{fig:gen_teaser}, while single-view reconstructions may appear plausible, they often fail to capture crucial geometric details necessary for manipulation~\cite{liu2025avractivevisiondrivenrobotic}.

% \vspace{-20pt}
To address these challenges, we develop an active exploration strategy. We first employ Grounded-SAM~\cite{ren2024grounded} to detect object-bounding boxes and masks from a global camera perspective. Combined with depth information, this provides initial object point clouds and spatial coordinates. The robot arm then systematically captures multi-perspective RGB observations $\mathcal{O}_{explore} \in \R^{C\times H\times W}$ using its wrist camera, where $C$ denotes the number of viewpoints. This exploration ensures comprehensive object coverage while accounting for potential occlusions during the subsequent manipulation phase.

\myparagraph{Object 3D Model Generation.}
\label{sec:digital-twin-generation}
% We employ arbitrary digital asset generation tools \footnote{We use Deemos's 3D digital asset Generation Model (from text or image) Rodin: \href{https://hyperhuman.deemos.com/rodin}{https://hyperhuman.deemos.com/rodin}} to reconstruct high-quality digital assets of target objects 
%
After obtaining multi-view observations, we utilize foundation model-based 3D asset generation~\cite{xu2023rodin} to reconstruct high-quality digital twins. This automated process leverages the model's embedded knowledge about common objects to accurately complete even partially visible regions. When faced with occlusions, such as a mug handle hidden from certain views, the model's prior knowledge enables plausible reconstruction of these unseen parts, providing a complete object representation crucial for subsequent manipulation planning.
To ensure reconstruction quality, the generated digital twins are evaluated against the observed views for geometric and textural consistency. This verification step helps maintain the fidelity of downstream semantic understanding. The reconstructed twins serve dual purposes: providing a basis for comprehensive semantic feature extraction and enabling accurate pose tracking during dynamic interactions.

\myparagraph{Virtual Semantic Flow Generation.}
\label{sec:semantic-flow-gen}
The digital twins provide a crucial advantage in overcoming real-world sensing limitations. Real cameras often produce incomplete or noisy depth information, with many sensors having invalid regions or limited resolution. In contrast, our virtual space allows the generation of high-resolution RGBD observations from arbitrary viewpoints, enabling the creation of complete object representations unconstrained by physical sensing limitations.

Our semantic flow generation process begins with multi-perspective feature extraction. Multi-view RGB observations generated in virtual space are processed through DINOv2~\cite{oquab2023dinov2}, producing rich feature maps $O\in(C, H, W, D_{VFM})$ that capture both low-level geometric details and high-level semantic information crucial for manipulation. To enhance computational efficiency while preserving essential information, we employ PCA to compress these high-dimensional features to $D_{feat}$ dimensions. The PCA model is trained on virtual space features from the training dataset, ensuring stable and consistent feature extraction across different objects and viewpoints. This dimensionality reduction significantly improves the real-time performance of our system while maintaining semantic understanding.
Based on initial object poses obtained through spatial alignment, we arrange digital assets in the virtual space and synthesize complete semantic flows by combining multi-view features with precise virtual depth information. The resulting semantic flow is uniformly sampled to $K$ points using Farthest Point Sampling (FPS) to obtain $P_{init}$. This virtual space-based approach ensures accuracy independent of real-world observation noise and occlusions.

The generated semantic flow serves as a canonical representation that can be dynamically transformed during manipulation while maintaining semantic consistency. Since this flow is constructed in virtual space using complete object models, it remains robust to partial observations and occlusions that occur during real-world interactions. 
% More implementation details of this process can be found in the Appendix.

\begin{figure}[tb] 
    \centering
    \includegraphics[width=0.85\linewidth]{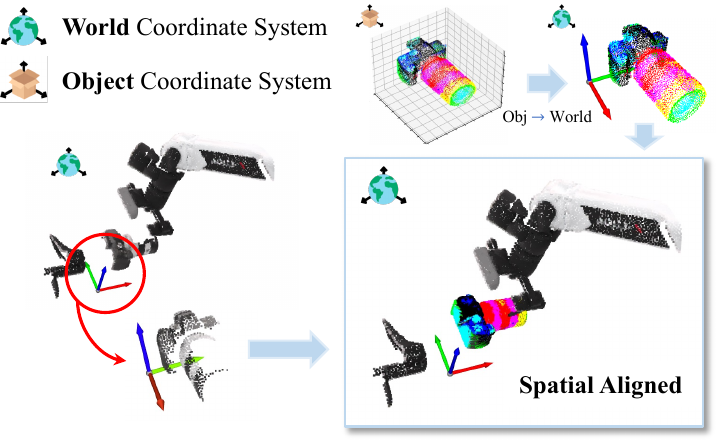}
    \vspace{-5pt}
    \caption{\textbf{Spatial alignment via object tracking.} We achieve alignment between the semantic flow and the physical object in real world by synchronizing the relative transformations of the object coordinate system to the world coordinate system.}
    
    \label{fig:spatial_alignment} 
    \vspace{-10pt}
\end{figure}

\subsection{Dynamic Semantic Flow Maintenance}
\myparagraph{Spatial Alignment via Object Tracking. }
\label{sec:spatial-alignment}
Once the initial semantic flow is established, maintaining its accuracy during dynamic manipulation becomes crucial. We achieve this through continuous spatial alignment between the semantic flow and the physical object. 

By integrating Grounded-SAM with task descriptions, we first detect and segment the target object from single-perspective RGB images to obtain masked RGBD observations. These observations, combined with the previously generated digital twin, enable FoundationPose~\cite{wen2024foundationpose} to compute the initial object pose matrix $M_{init}$.
During manipulation, we continuously update our pose estimates through FoundationPose, obtaining precise object poses $M_{update}$ at each timestep. This enables the dynamic transformation of the semantic flow through:
{\small
\vspace{-5pt}
\begin{equation}
P_{update}^T = [(M_{c2w}M_{update})(M_{c2w}M_{init})^{-1}]P_{init}^T.
\label{eq:1}
% \vspace{-3pt}
\end{equation}
}

The key advantage of our approach lies in FoundationPose's ability to maintain accurate pose estimation even under significant occlusions, leveraging the rich information contained in our digital twins. Since our feature point cloud is obtained from complete observations in virtual space, we consider it optimal. Rather than repeatedly detecting, segmenting, and computing features at each timestep—which could lead to compounding errors—we transform this high-quality feature point cloud directly. This approach not only provides accurate and complete semantic flow estimates during occlusions but also ensures computational efficiency and robustness during dynamic interactions.

% \vspace{20pt}
\subsection{\textbf{\ours-Enhanced Diffusion Policy}}
\label{sec:decision-making}
To effectively leverage our semantic flow for precise manipulation, we enhance diffusion policies through three key components: conditional feature acquisition, a conditional denoising process, and a specialized training procedure.

\begin{figure}[tb] 
    \vspace{-3pt}
    \centering
    \includegraphics[width=0.95\linewidth]{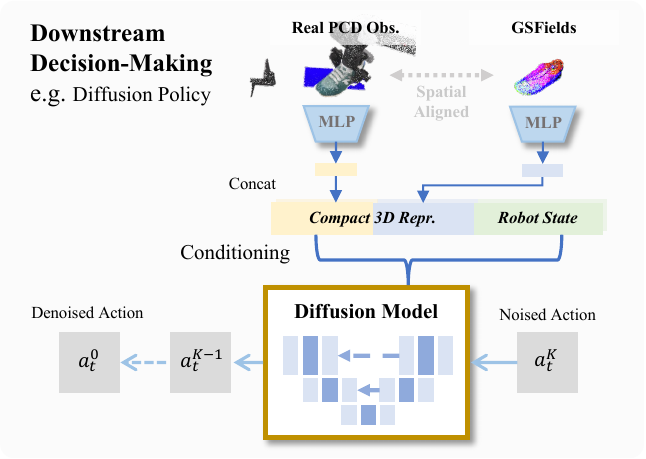}
    \vspace{-8pt}
    \caption{\textbf{\ours-enhanced diffusion policy.}}
    \label{fig:decision-making} 
    \vspace{-12pt}
\end{figure}

\myparagraph{Conditional Feature Acquisition.} 
Our policy integrates three distinct types of information through separate MLP encoders. First, the transformed and updated semantic flow with shape $(K, 3+D_{feat})$ is processed to obtain semantic features $f_s$, capturing rich object-centric semantic understanding. Second, the real point cloud observations with shape $(K, 3)$ are encoded to produce scene features $f_r$, providing immediate geometric feedback. Finally, the current robot joint states are encoded into robot state features $f_p$, ensuring awareness of the manipulator's configuration. This multi-modal feature acquisition enables our policy to reason about both semantic and geometric aspects of the task.

\myparagraph{Conditional Denoising Process.} 
We formulate the decision-making process as a conditional denoising operation, where actions are generated by gradually denoising random Gaussian noise conditioned on our extracted features. Beginning with random noise $a^K$, a denoising network $\epsilon_{\theta}$ performs $K$ iterations to predict the final action $a^0$:
{\small
\vspace{-8pt}
\begin{equation}
    a^{k-1} = \alpha_k (a^k-\gamma_k, k, \epsilon_\theta(a^k,f_{s},f_{r},f_{p}))+\sigma_k\mathcal{N}(0,\mathbf{I}),
\label{eq:2}
\vspace{-2pt}
\end{equation}
}
where $\alpha_k$, $\gamma_k$ and $\sigma_k$ are functions of the denoising step $k$ and depend on the noise scheduler. This formulation allows our policy to capture complex action distributions while maintaining stability through structured denoising process.

\myparagraph{Training Procedure.} 
We employ the DDIM scheduler for noise scheduling and optimize a noise prediction objective. The training loss is formulated as:
{\small
\vspace{-5pt}
\begin{equation}
    \mathcal{L}=\text{MSE}\left(\boldsymbol{\epsilon}^k, \boldsymbol{\epsilon}_\theta(\bar{\alpha_k} a^0 + \bar{\beta_k} \boldsymbol{\epsilon}^k, k, f_s, f_r, f_p)\right).
\label{eq:3}
\vspace{-6pt}
\end{equation}
}

This loss function trains the network to predict the noise added to the expert actions, enabling effective learning from demonstration data. The inclusion of semantic flow features $f_s$ alongside real observations $f_r$ and robot state $f_p$ allows the policy to leverage both geometric precision and semantic understanding during execution.

\definecolor{newgreen}{RGB}{78, 173, 102}
\newcommand{\xmark}{\textcolor{red}{\ding{55}}} % Red cross

\begin{table*}[tbp]
\small
  \centering
    \resizebox{0.8\linewidth}{!}{
    \begin{tabular}{l|cccc|c}
    \toprule
          & \textbf{\textit{Shoe Place (T)}} & \textbf{\textit{Dual Shoes Place (T)}} & \textbf{\textit{Tool Adjust (T)}} & \textbf{\textit{Bottle Adjust (T)}} & \textbf{\textit{Average}} \\
    \midrule

    % \ours w/o full  &  &  & \\
    DP & $26.0\pm{11.4}$ & $3.3\pm{1.2}$  & $21.7\pm{5.1}$ & $16.3\pm{4.6}$  & 16.8 (\textcolor{purple}{ $\downarrow$51.5}) \\
    3D DP  & $54.0\pm{6.9}$ & $13.0\pm{1.7}$  &   $43.3\pm{9.1}$  & $74.3\pm{8.6}$  & 46.2 (\textcolor{purple} {$\downarrow$22.1})\\
    3D DP w/ color & $58.0\pm{3.0}$  & $7.0\pm{1.0}$  &  $70.3\pm{9.3}$  & $41.0\pm{12.8}$  & 44.1 (\textcolor{purple} {$\downarrow$24.2}) \\
    \textbf{DP w/ \ours} & $\textbf{83.0}\pm{3.6}$\cellcolor{table_color} & $\textbf{24.0}\pm{3.6}$\cellcolor{table_color} & $\textbf{84.3}\pm{10.1}$\cellcolor{table_color} & $\textbf{82.0}\pm{4.0}$ \cellcolor{table_color}& \textbf{68.3}\hspace{1.0cm} \cellcolor{table_color}\\
    % ACT & & & \\

    \bottomrule
    \end{tabular}}
    \vspace{-7pt}
    \caption{\textbf{Success rates (in \%) of simulation tasks for terminal constraint control tasks.} We report the mean and standard deviation computed over 3 random seeds.}
    \vspace{-5pt}
  \label{tab:terminal_constraint}%
\end{table*}

\begin{table*}[htbp]
\small
  \centering
  \resizebox{0.8\linewidth}{!}{
    \begin{tabular}{l|cccc|c}
    \toprule
          & \textbf{\textit{Shoe Place (G)}} & \textbf{\textit{Dual Shoes Place (G)}} & \textbf{\textit{Diverse Bottles Pick (G)}} & \textbf{\textit{Tool Adjust (G)}}  & \textbf{\textit{Average}} \\
    \midrule

    % \ours w/o full  &  &  & \\
    DP & $17.7\pm{3.2}$ & $3.0\pm{2.6}$  & $8.7\pm{3.2}$ & $16.0\pm{11.3}$  & 11.4 (\textcolor{purple} {$\downarrow$38.7}) \\
    3D DP  & $51.0\pm{6.6}$ & $9.3\pm{3.1}$  & $39.3\pm{9.6}$  & $27.0\pm{11.5}$ & 31.7 (\textcolor{purple} {$\downarrow$18.4}) \\
    3D DP w/ color & $38.7\pm{7.5}$ & $8.0\pm{1.7}$  &  $13.0\pm{5.0}$ & $57.3\pm{3.5}$ & 29.3 (\textcolor{purple} {$\downarrow$20.8})\\
    \textbf{DP w/ \ours} & $\textbf{63.7}\pm{3.5}$\cellcolor{table_color} & $\textbf{14.7}\pm{2.1}$ \cellcolor{table_color} & $\textbf{51.3}\pm{10.4}$\cellcolor{table_color} & $\textbf{70.7}\pm{11.7}$\cellcolor{table_color} & \textbf{50.1}\hspace{1.0cm} \cellcolor{table_color}\\

    \bottomrule
    \end{tabular}}
    \vspace{-8pt}
    \caption{\textbf{Success rates (in \%) of cross-object generalization tasks.}. We report the mean and standard deviation computed over 3 random seeds.}
    \vspace{-12pt}
  \label{tab:generalization}%
\end{table*}

% \begin{table}[htbp]
% \small
%   \centering
%     \caption{Sim-to-Real Tasks}
%     \vspace{-5pt}
%     \resizebox{\linewidth}{!}{
%     \begin{tabular}{lccc}
%     \toprule
%           & \textbf{\textit{Bottle Pick}} & \textbf{\textit{Bottle Pick Hard}} & \textbf{\textit{Diverse Bottles Pick}} \\
%     \midrule
%     \textbf{\ours} &  & &  \\
%     3D DP  &  &   & \\
%     \bottomrule
%     \end{tabular}}%
%   \label{tab:sim2real}%
% \end{table}

% \begin{table*}[tbp]
% \small
%   \centering
%     \caption{Sim2Real Task Result.}
%     \begin{tabular}{lccc}
%     \toprule
%           & \textbf{\textit{Shoe Place}} & \textbf{\textit{Shoes Place}} & \textbf{\textit{Diverse Bottles Pick}} \\
%     \midrule

%     \textbf{\ours} &  & &  \\
%     3D DP  &  &   & \\

%     \midrule
%               & \textbf{\textit{Tool Pick}} & \textbf{\textit{Bottle Adjust}} & \textbf{\textit{Average}} \\
%     \midrule

%     \textbf{\ours}  &  & &  \\
%     3D Diffusion Policy  &  &  & \\
%     \bottomrule
%     \end{tabular}%
%   \label{tab:main-results}%
% \end{table*}

\section{Experiments}
\label{sec:exp}

We conduct extensive experiments to evaluate \ours's effectiveness across two key aspects: terminal constraint satisfaction and cross-object generalization.
% and 2) whether \ours can serve as an effective observational bridge between simulation and reality.

\subsection{Experimental Setup}

We evaluate our approach on five distinct manipulation tasks from the RoboTwin benchmark~\cite{mu2024robotwin}, as illustrated in Figure \ref{fig:sim_task}. Each task is designed to assess specific aspects, detailed task descriptions can be found in Appendix~\ref{appendix:tasks}.

\begin{figure}[tb] 
    % \hspace{-10pt}
    \centering
    \includegraphics[width=0.95\linewidth]{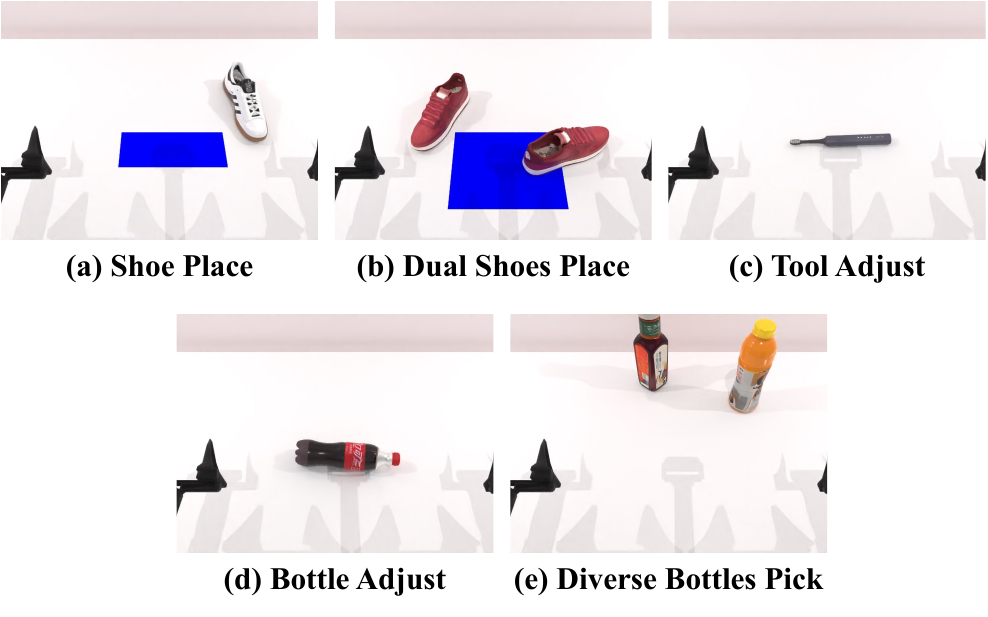}
    \vspace{-10pt}
    \caption{\textbf{Five testing benchmark tasks.}}
    \vspace{-18pt}
    \label{fig:sim_task} 
\end{figure}

\myparagraph{Single-arm Tasks.}
(1) \textbf{Shoe Place:} Position a randomly placed shoe on a mat with toe facing left;
(2) \textbf{Bottle Adjust:} Grasp bottles based on opening orientation, ensuring upright placement;
(3) \textbf{Tool Adjust:} Manipulate various tools by their handles based on head orientation.

\myparagraph{Dual-arm Tasks:}
(1) \textbf{Dual Shoes Place:} Position two shoes with synchronized bi-manual control,
(2) \textbf{Diverse Bottles Pick:} Handle bottles of varying sizes using arms.

% \myparagraph{Shoe Place:} The robot is required to place a single, randomly positioned shoe on a blue mat with the toe facing left. This task involves generalization to unseen shoes, evaluating intra-class generalization, and adherence to terminal constraints.

% \myparagraph{Dual Shoes Place}: The robot must position two randomly placed shoes on a blue mat with both toes facing left, using both arms. This task also requires generalization to unseen shoes, assessing intra-class generalization, and terminal constraint satisfaction.

% \myparagraph{Bottle Adjust}: Randomly placed bottles on a table must be picked up by the robot arm. The right arm is used if the bottle’s opening faces left, and the left arm if it faces right, ensuring that the bottle is upright when lifted. This setup evaluates intra-class generalization and terminal constraint satisfaction for unseen bottles.

% \myparagraph{Tool Adjust}: Randomly positioned tools with similar handles are placed on a table. The robot arm must grip the handle using the right arm if the tool’s head faces left, and the left arm if it faces right, ensuring the head points upward after lifting. This task requires generalization to semantically similar but unseen objects, assessing cross-category generalization and terminal constraint satisfaction.

% \myparagraph{Diverse Bottles Pick}: Two different-sized bottles are randomly placed on a table, and the robot must learn to use both arms to handle bottles of varying shapes and sizes. This task assesses intra-class generalization to novel bottles.

For each task, we train policies using 100 expert demonstrations and evaluate across 3 random seeds with 100 test episodes per seed. To assess different capabilities, we maintain separate object sets for terminal constraint and generalization evaluations. Performance is measured through average success rates and standard deviations across seeds. We employ PCA to reduce the feature dimensions of DINOv2 to 5, and downsample both the original point cloud and the virtual point cloud to 1024 points.

% \subsubsection{Baselines}
\myparagraph{Baselines:}
We use the 3D Diffusion Policy (DP3)~\cite{ze20243d}, which utilizes efficient point encoders to create compact 3D representations, and its variant with RGB color information DP3(w/ color), as well as the 2D Diffusion Policy (DP)~\cite{chi2023diffusion} processing visual information in images to predict robot actions, as key baselines.
% To explore the enhancing effect of \ours as a semantic supplement for policies to achieve pose-aware requirements and generalization tasks, we use imitation learning algorithms DP3 (w/ color, w/o color) and DP which rely solely on traditional point cloud and RGB observations as baselines.
%
% We use the 3D Diffusion Policy (DP3)\cite{ze20243d}, which utilizes efficient point encoders to create compact 3D representations for imitation learning, and the 2D Diffusion Policy (DP)~\cite{chi2023diffusion}, which processes visual information in images to predict actions for robotic manipulation tasks, as key baselines.
% addresses these limitations by incorporating three-dimensional visual representations through point clouds. By using efficient point encoders to create compact 3D representations, DP3 enhances spatial awareness and demonstrates improved performance in tasks requiring complex spatial understanding.
%
We train 3000 epochs for all the tasks with batch size 256 for \ours and DP3. For DP, we train 300 epochs for all the tasks with batch size 128.

% \subsubsection{Sim-to-Real Tasks}
% \todo{sim2real}

% \subsection{Results and Analysis}
% We analyze our experimental results from three key perspectives: terminal constraint satisfaction, generalization capability, and ablation study. The visualization of \ours in the five simulation tasks is shown in Figure~\ref{fig:flow_vis}.

\begin{figure}[tb] 
    \centering
    \includegraphics[width=0.45\textwidth]{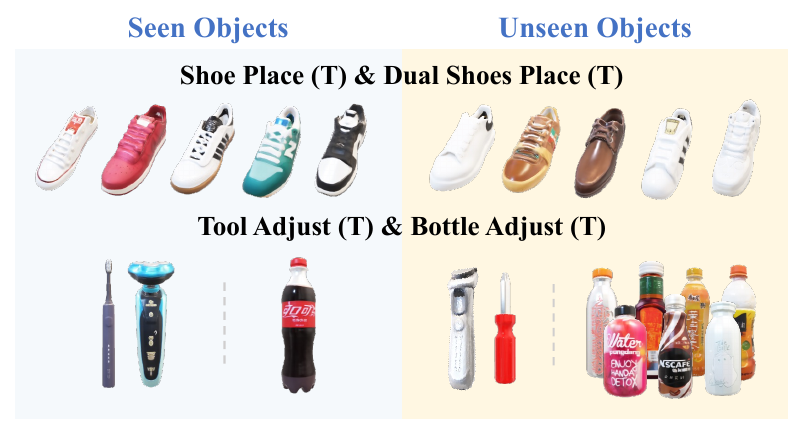}
    \vspace{-8pt}
    \caption{\textbf{Seen and unseen object sets for four tasks with high terminal constraint requirements.}}
    \vspace{-12pt}
    \label{fig:task_for_terminal_constraint} 
\end{figure}

\subsection{Evaluation on Pose-aware Manipulation Tasks}

To investigate the ability of our method to provide semantic information that enhances the policy's understanding of the semantics of the manipulated object parts, we selected \textit{Shoe Place}, \textit{Dual Shoes Place}, \textit{Tool Adjust}, and \textit{Bottle Adjust} as test tasks, requiring the robotic arm to meet pose-aware requirements. We chose objects that are geometrically similar to the training set for testing, to reduce the examination of the model's generalization ability, as shown in Fig.~\ref{fig:task_for_terminal_constraint}. We chose unseen objects as the test set to avoid the situation that the policy memorizes training objects, which cheats the performance results.

As shown in Tab.~\ref{tab:terminal_constraint}, \ours consistently outperforms baseline methods in achieving pose-aware requirements across all four tasks. Our method achieves over 25\% higher success rates in the \textit{Shoe Place (T)} task for correct orientation and in the \textit{Bottle Adjust (T)} task, we achieved a success rate that exceeded the average of the baselines by over 38\% for upright pick. This demonstrates that the semantic understanding provided by \ours helps the policy better comprehend and execute pose-aware requirements.

The performance gain is particularly notable in tasks requiring precise object orientation, such as \textit{Dual Shoes Place (T)}. While baseline methods occasionally achieve correct positioning, they struggle with maintaining consistent orientation accuracy. \ours nearly doubles the success rate compared to the strongest baseline, suggesting that our semantic representations effectively encode spatial relationships and object orientations.

\subsection{Evaluation on Generalization Performance}
\begin{figure}[tb] 
    \centering
    \includegraphics[width=0.45\textwidth]{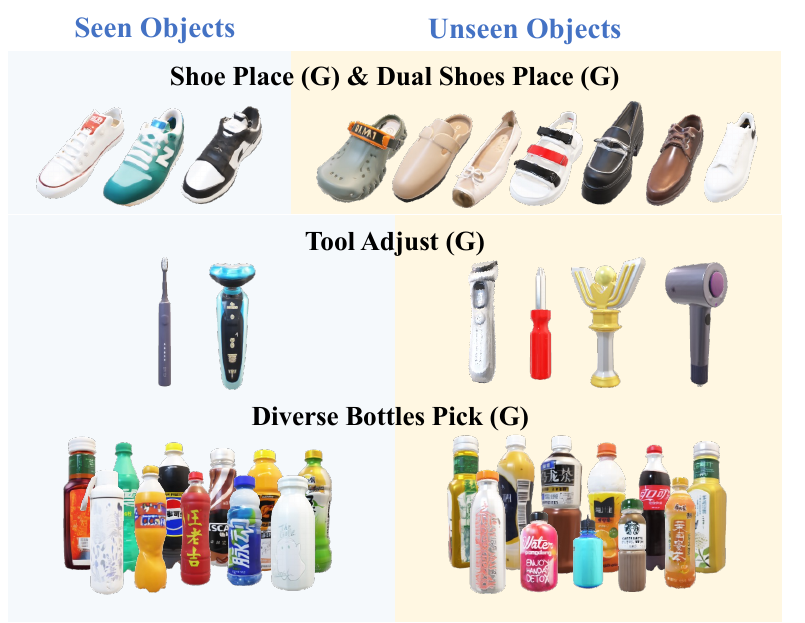}
    \vspace{-8pt}
    \caption{\textbf{Cross-object generalization settings.} Seen and unseen object sets for four tasks with high generalization requirements.}
    \vspace{-12pt}
    \label{fig:task_for_generalization} 
\end{figure}
To investigate the generalization capability of our method in providing semantic information for manipulating objects, we have selected \textit{Shoe Place}, \textit{Dual Shoes Place}, \textit{Tool Adjust}, and \textit{Diverse Bottles Pick} as test tasks. Unlike tasks that require the satisfaction of terminal constraints, we choose as few and similar visible objects as possible for the training set and select objects that are as geometrically distinct as possible from the training set for the test set, as shown in Figure~\ref{fig:task_for_generalization}. This requires the policy to correctly manipulate objects that are geometrically different from those it has seen with only a limited exposure, focusing on assessing the policy's generalization ability.

Our method achieves an average success rate 18.4\% higher than strongest baseline across four tasks, exhibiting powerful generalization capabilities across different object categories and variations, as shown in Table~\ref{tab:generalization}:

\begin{itemize}
    \item \textbf{Intra-class Generalization}: In tasks involving geometrically distinct unseen instances of the same object category (\textit{Shoe Place (G)}, \textit{Dual Shoes Place (G)}, \textit{Diverse Bottles Pick (G)}), our method maintains optimal performance, indicating that \ours encompasses a genuine semantic understanding of objects, enabling effective operation generalization even when faced with geometrically diverse instances within the same category.
    
    \item \textbf{Cross-category Generalization}: For the \textit{Tool Adjust (G)} task, which necessitates dealing with objects that are semantically similar but belong to different categories, our method must learn to grasp positions akin to a handle on the objects while also fulfilling the pick-upright condition. \ours achieved a success rate of \textbf{70.7\%} on previously unseen tool categories, which is \textbf{13.4\%} higher than the best baseline. This result confirms the capability of our method to transfer learned operational skills across different object categories.

    \item \textbf{Scale Variation}: In the \textit{Diverse Bottles Pick (G)} task, \ours successfully generalizes to bottles of varying sizes, maintaining a consistent grasp success rate of \textbf{51.3\%}. This indicates robust handling of geometric variations while preserving semantic understanding.
\end{itemize}

\begin{figure*}[h]
    \centering
    \includegraphics[width=1.0\linewidth]{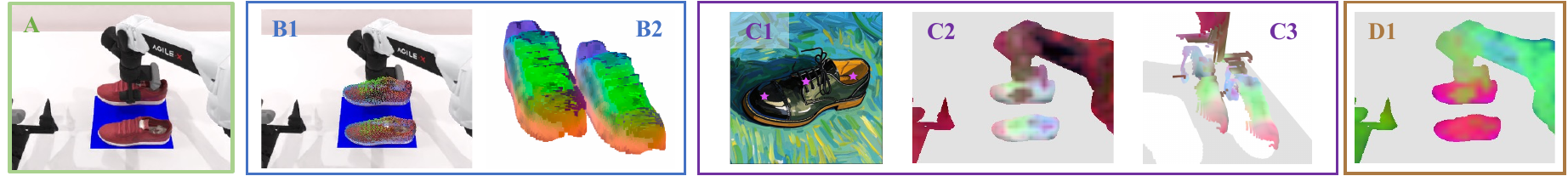}
    \vspace{-19pt}
    \caption{\textbf{Feature Quality Visualization.} \textbf{\textcolor{Acolor}{A}}: raw RGB, \textbf{\textcolor{Bcolor}{B}}: G3Flow, \textbf{\textcolor{Ccolor}{C} and \textbf{\textcolor{Dcolor}{D}}}: Scene-level DINOv2 feature.}
    \vspace{-13pt}
    \label{fig:vis_d3fields} 
\end{figure*}

\begin{figure*}[tb] 
    \centering
    \includegraphics[width=0.95\textwidth]{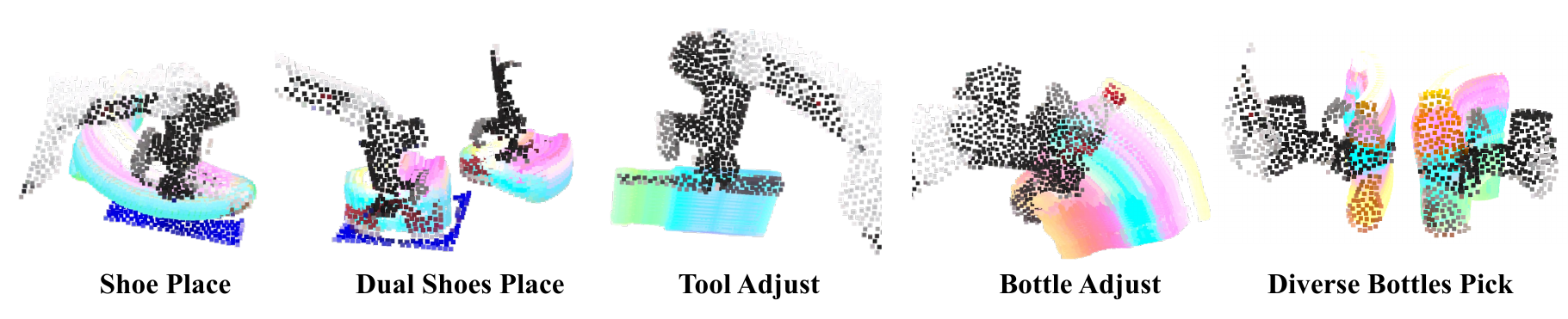}
    \vspace{-8pt}
    \caption{\textbf{Visualization of \ours in the 5 evaluation tasks.}}
    \vspace{-9pt}
    \label{fig:flow_vis} 
\end{figure*}

\subsection{Ablation Study}

\begin{table}[tb]
\small
  \centering
  \resizebox{1.0\linewidth}{!}{
    \begin{tabular}{l|cc}
    \toprule
        \textbf{Method}  &  \textbf{Field-Gen Freq.} &  \textbf{Decision-making Freq.}\\
    \midrule
    DP w/ Scene-Level DINOv2 & 10.52 Hz  (\textcolor{purple} {$\downarrow$75.5\%}) & 9.52 Hz  (\textcolor{purple} {$\downarrow$72.0\%}) \\
    DP w/ D$^3$Fields (GenDP) & 8.14 Hz  (\textcolor{purple} {$\downarrow$81.0\%}) &  6.89 Hz  (\textcolor{purple} {$\downarrow$79.8\%}) \\
    DP w/ \alias & \bestcell{\textbf{42.92 Hz}} & \bestcell{\textbf{34.04 Hz}} \\
    \bottomrule
    \end{tabular}}
    \vspace{-8pt}
    \caption{\textbf{Comparison of Computational Efficiency.}}
    \vspace{-12pt}
  \label{tab:time}%
\end{table}

\myparagraph{Ablation on Efficiency.}
Robotic manipulation tasks have stringent requirements for real-time performance. We test the model inference speed on a single NVIDIA GeForce RTX 4090. The results are shown in Tab.~\ref{tab:time}. Our method significantly outperforms baselines, achieving a decision frequency of 34.04 Hz, nearly 6 times faster than GenDP~\cite{wang2024gendp}, meeting the requirements of most real-time robotics manipulation tasks.

Our pipeline generates digital assets once per object, even across multiple operations. After the initial DINOv2 feature extraction, subsequent semantic flow estimation relies on a lightweight pose-tracking network~\cite{wen2024foundationpose, wang2024articulated}, enabling real-time performance.

\myparagraph{Ablation on Quality of Semantic Field.}
% To explore the advantages of our complete, dynamic, object-level semantic flow representation, 

We conducted an ablation study comparing our method against conventional scene-level feature clouds and D$^3$Fields, using the \textit{Shoe Place (T)} and \textit{Dual Shoes Place (T)} tasks, which require long-term semantic reasoning and involve occlusions. As shown in Table~\ref{tab:ablation_1}, our approach improves success rates by 22.6\% and 41.2\% over scene-level features, and by 9.3\% and 3.7\% over D$^3$Fields. While D$^3$Fields benefits from human prior knowledge, our method outperforms it by focusing on object-centered visual inputs, which reduces irrelevant background noise (Sec.~\ref{sec:object-centric-exploration}). Moreover, scene-level baselines struggle with occlusions, whereas our tracking-based semantic flow maintains robust object representations even when occluded, enhancing feature quality and task performance.

% We conduct the ablation study with comparisons to conventional scene-level feature clouds and D$^3$Fields. 
% %
% We selected the \textit{Shoe Place (T)} and \textit{Dual Shoes Place (T)} tasks as a testbed because they require adjustments of the shoe orientation throughout the trajectory, relying more on long-term semantic understanding. Additionally, object occlusions are designed into the tasks, posing a greater challenge for semantic comprehension.

% As shown in Table~\ref{tab:ablation_1}, our approach achieves a 22.6\% and 41.2\% increase in success rates for scene-level features in the \textit{Shoe Place (T)} and \textit{Dual Shoes Place (T)} tasks, respectively. For D$^3$Fields features, the success rates improve by 9.3\% and 3.7\%, even though D$^3$Fields outperforms scene-level feature clouds, as it incorporates human prior knowledge, making the features more effective. \ours performs better as the images we send into the VFM are centered around the object, filtering out unrelated information, as detailed in Sec.~\ref{sec:object-centric-exploration}. 
% Additionally, the baselines' scene-level feature clouds cannot handle occlusions that often occur during object interactions. However, \ours is able to obtain complete object semantic flow through tracking and transformation method, even under occlusion, leading to an enhancement in feature quality.

\begin{table}[tbp]
\small
  \centering
  \resizebox{0.99\linewidth}{!}{
    \begin{tabular}{l|cc}
    \toprule
          & \textbf{Shoe Place (T)} & \textbf{Dual Shoes Place (T)}\\
    \midrule
    DP w/ Scene-Level Feature & $67.7\pm{1.5}$ & $17.0\pm{1.7}$ \\
    DP w/ D$^3$Fields (GenDP) & $73.7\pm{2.5}$ & $20.3\pm{6.8}$ \\
    DP w/ \ours  & \bestcell{$\textbf{83.0}\pm{3.6}$} & \bestcell{$\textbf{24.0}\pm{3.6}$}  \\
    \bottomrule
    \end{tabular}}
    \vspace{-7pt}
    \caption{\textbf{Ablation on Quality of Semantic Field.} We compare the success rates of scene-level features, D$^3$Fields and \ours on \textit{Shoe Place} and \textit{Dual Shoes Place} tasks.}
  \label{tab:ablation_1}%
  \vspace{-12pt}
\end{table}

In Fig.~\ref{fig:vis_d3fields}, we compare the feature quality of our method with that of \textbf{\textcolor{Dcolor}{scene-level feature}} and \textbf{\textcolor{Ccolor}{D$^3$Fields}}. It can be observed that due to occlusions and background interference, scene-level features (\textbf{\textcolor{Dcolor}{D1}}) fail to distinguish the shoe's toe and heel, which is crucial for adjusting the shoe's pose. D$^3$Fields benefits from goal image priors (\textbf{\textcolor{Ccolor}{C1}}), which improves semantic preservation over raw DINOv2. However, the occluded regions (\eg, gripper coverage, \textbf{\textcolor{Ccolor}{C2}}) of it suffer from inaccurate semantics, and it cannot recover a complete semantic point cloud from a single view, limiting precise manipulation (\textbf{\textcolor{Ccolor}{C3}}). In contrast, \ours generates a complete and high-quality semantic field by digital twin (\textbf{\textcolor{Bcolor}{B2}}).

\subsection{Ablation on VFMs.}
% \vspace{-10pt}

\begin{table}[htbp]
\small
  \centering
  \resizebox{0.99\linewidth}{!}{
    \begin{tabular}{l|ccc}
    \toprule
       Method   & \textbf{G3Flow (w/ CLIP)} & \textbf{G3FLow (w/ SAM)} & \textbf{G3Flow}\\
    \midrule
    Success Rate & $72.3\pm{2.4}$ & $74.7\pm{5.0}$ & \bestcell{$83.0\pm{3.6}$} \\
    \bottomrule
    \end{tabular}}
    \vspace{-7pt}
    \caption{\textbf{Ablation on VFMs.} Success rates of G3Flow implemented with different VFMs (our method uses DINOv2) on the \textit{Shoe Place (T)} task.}
  \label{tab:ablation_vfms}%
  \vspace{-12pt}
\end{table}

We conducted an ablation study on VFM replacement in the \textit{Shoe Place (T)} task. As shown in Tab.~\ref{tab:ablation_vfms}, our DINOv2 outperforms CLIP and SAM in pose awareness and generalization, as CLIP prioritizes image-text alignment over spatial understanding, while SAM focuses excessively on detailed features rather than part-level characteristics. This highlights the importance of selecting appropriate features for effective policy learning.

\subsection{Visualizations of G3Flow}
The visualization of \ours across our five evaluation tasks is shown in Fig.~\ref{fig:flow_vis}, demonstrating how our real-time semantic flow maintains both temporal coherence and spatial alignment during manipulation. In each task visualization, different colors represent distinct semantic features: orientation-critical regions (pink) for shoe placement tasks, functional parts (blue/green) for tool and bottle manipulation, and consistent semantic representations across varied object sizes in the diverse bottles task. 
Notably, our semantic flow remains complete and stable even under partial occlusions from robot arms or objects, validating the effectiveness of our foundation model-driven approach.

\section{Conclusion}
In this paper, we introduced \ours, a novel framework that leverages foundation models to construct real-time semantic flow for enhanced robotic manipulation. 
%
% Our approach addresses key limitations in existing geometric-centric methods through semantic flow, a dynamic, object-centric 3D semantic representation maintained throughout manipulation processes. 
%
By uniquely integrating 3D generative models for digital twin creation, vision foundation models for semantic feature extraction, and robust pose tracking, \ours enables complete semantic understanding while eliminating manual annotation requirements.
Extensive experiments demonstrate \ours's effectiveness in both terminal-constrained manipulation and cross-object generalization tasks. 
They validate our key insight that maintaining consistent semantic understanding through foundation model integration can substantially improve manipulation performance, particularly in scenarios requiring precise control and object variation handling.
Looking forward, our method establishes a foundation for semantic - aware robotic manipulation. Future research directions include articulated - object and multi - object interactions, as well as continuous optimization of 3D modeling for interacted objects.

\clearpage
\section*{Acknowledgements}
This paper is partially supported by the National Key R\&D Program of China No.2022ZD0161000, the General Research Fund of Hong Kong No.17200622 and 17209324, and the Jockey Club STEM Lab of Autonomous Intelligent Systems funded by The Hong Kong Jockey Club Charities Trust. 
{
    \small
    \bibliographystyle{ieeenat_fullname}
    \bibliography{main}
}

\clearpage
\onecolumn
\appendix
\setcounter{page}{1}
{\centering
\Large
\textbf{\thetitle}\\
\vspace{0.3em}Supplementary Material \\
\vspace{0.5em}
}

\section{Simulation Tasks}
\label{appendix:tasks}
We provide detailed descriptions of all simulation tasks, as shown in Table \ref{tab:benchmark_description}, totaling 5 tasks.
\begin{table*}[h]
\centering
\small

\resizebox{0.75\textwidth}{!}{
\begin{tabular}{>{\itshape\centering}m{3.5cm} p{10cm}} 
\toprule
\textbf{Task}   & \textbf{Description}        \\ \midrule 
Bottle Adjust & A bottle is placed horizontally on the table. The bottle's design is random and does not repeat in the training and testing sets. When the bottle's head is facing left, pick up the bottle with the right robot arm so that the bottle's head is facing up; otherwise, do the opposite. \\ \midrule
Tool Adjust & A tool is placed horizontally on the table. The tool's design is random and does not repeat in the training and testing sets. When the tool's head is facing left, pick up the tool with the right robot arm so that the tool's head is facing up; otherwise, do the opposite. \\ \midrule
Diverse Bottles Pick & A random bottle is placed on the left and right sides of the table. The bottles' designs are random and do not repeat in the training and testing sets. Both left and right arms are used to lift the two bottles to a designated location. \\ \midrule
Shoe Place & Shoes are placed randomly on the table, with random designs that do not repeat in the training and testing sets. The robotic arm moves the shoes to a blue area in the center of the table, with the shoe head facing the left side of the table. \\ \midrule
Dual Shoes Place & One shoe is placed randomly on the left and right sides of the table. The shoes are the same pair with random designs that do not repeat in the training and testing sets. Both left and right arms are used to pick up the shoes and place them in the blue area, with the shoe heads facing the left side of the table. \\ \bottomrule
\end{tabular}}
\caption{Benchmark Task Descriptions.}
\vspace{-5pt}
% \scriptsize
\label{tab:benchmark_description}
\end{table*}

\section{Implementation Details}

This section will provide a detailed introduction to the implementation details of G3Flow as described in the paper, including the setup of the experiments. 

\subsection{Structure Details}
\myparagraph{Vision Foundation Model.} We utilize the ViT-S/14 variant and transform all images to a resolution of 420 by 420 pixels. These are then fed into the model to obtain feature maps of size 30 by 30, where each pixel has a 384-dimensional feature representation. Subsequently, these features are transformed back to the original image dimensions. The PyTorch implementation is as follow:
\begin{lstlisting}[language=Python, frame=none, basicstyle=\small\ttfamily, commentstyle=\color{blue}\small\ttfamily,columns=fullflexible, breaklines=true, postbreak=\mbox{\textcolor{red}{$\hookrightarrow$}\space}, escapeinside={(*}{*)}]
def get_dino_feature(image, transform_size=420, model=None):
    img, H, W = transform_np_image_to_torch(image, transform_size=transform_size) 
    res = model(img) # torch.Size([1, 384, 30, 30])
    feature = np.array(res.cpu().unsqueeze(0))
    new_order = (0, 1, 3, 4, 2) # torch.Size([1, 30, 30, 384])
    feature = np.transpose(feature, new_order)
    orig_shape_feature = transform_shape(torch.Tensor(np.transpose(feature[0], (0, 3, 1, 2))), H, W)
    orig_shape_feature_line = orig_shape_feature.reshape(-1, orig_shape_feature.shape[-1])
    return orig_shape_feature, orig_shape_feature_line
\end{lstlisting}

\myparagraph{PCA.} We employ Principal Component Analysis (PCA) to reduce the feature dimensionality from 384 to 5.

\myparagraph{Perception.} For image observations, we uniformly employ a camera setup with a resolution of 320 by 240 pixels and a field of view (fovy) of 45 degrees. We apply Farthest Point Sampling (FPS) to both the feature point cloud and the real observation point cloud, downsampling them to 1024 points. We provide a simple PyTorch implementation of our Feature Pointcloud Encoder as follows:
\begin{lstlisting}[language=Python, frame=none, basicstyle=\small\ttfamily, commentstyle=\color{blue}\small\ttfamily,columns=fullflexible, breaklines=true, postbreak=\mbox{\textcolor{red}{$\hookrightarrow$}\space}, escapeinside={(*}{*)}]
class PointNetFeaturePCDEncoder(nn.Module):
    def __init__(self,
                 in_channels,
                 out_channels,
                 use_layernorm,
                 final_norm,
                 use_projection,
                 **kwargs
                 ):
        super().__init__()
        block_channel = [512, 512, 256]
       
        self.mlp = nn.Sequential(
            nn.Linear(in_channels, block_channel[0]),
            nn.LayerNorm(block_channel[0]) if use_layernorm else nn.Identity(),
            nn.ReLU(),
            nn.Linear(block_channel[0], block_channel[1]),
            nn.LayerNorm(block_channel[1]) if use_layernorm else nn.Identity(),
            nn.ReLU(),
            nn.Linear(block_channel[1], block_channel[2]),
            nn.LayerNorm(block_channel[2]) if use_layernorm else nn.Identity(),
        )
        
        self.final_projection = nn.Sequential(
            nn.Linear(block_channel[-1], out_channels),
            nn.LayerNorm(out_channels)
        )

        self.use_projection = use_projection
         
    def forward(self, x):
        x = self.mlp(x)
        x = torch.max(x, 1)[0]
        x = self.final_projection(x)
        return x
\end{lstlisting}

\newpage
\subsection{Parameter Details}
\myparagraph{Training Setup.} The training setup for the Diffusion Policy based on G3Flow is shown in Tab.~\ref{tab:G3Flow_setup}.
\begin{table*}[h!]
\centering
\small
\begin{tabular}{cc}
\midrule
Parameter & Value \\ 
\midrule
horizon & 8 \\ 
n\_obs\_steps & 3 \\ 
n\_action\_steps & 6 \\ 
num\_inference\_steps & 10 \\ 
dataloader.batch\_size & 256 \\ 
dataloader.num\_workers & 8 \\ 
dataloader.shuffle & True \\ 
dataloader.pin\_memory & True \\ 
dataloader.persistent\_workers & False \\ 
optimizer.\_target\_ & torch.optim.AdamW \\ 
optimizer.lr & 1.0e-4 \\ 
optimizer.betas & [0.95, 0.999] \\ 
optimizer.eps & 1.0e-8 \\ 
optimizer.weight\_decay & 1.0e-6 \\ 
training.lr\_scheduler & cosine \\ 
training.lr\_warmup\_steps & 500 \\ 
training.num\_epochs & 3000 \\ 
training.gradient\_accumulate\_every & 1 \\ 
\bottomrule
\end{tabular}
\vspace{-5pt}
\caption{Model Training Settings. Hyper-parameter Settings for Training and Deployment of G3Flow-empowered DP.}
\label{tab:G3Flow_setup}
\vspace{-10pt}
\end{table*}

\myparagraph{Baselines Setup.} We outline the key training settings for the baseline in Tab.~\ref{tab:baseline_setup}.

\begin{table*}[h!]
% \vspace{-40em}
\small
\centering
\begin{tabular}{ccc}
\midrule
Parameter & DP & DP3 \\ 
\midrule
horizon & 8 & 8 \\ 
n\_obs\_steps & 3 & 3 \\ 
n\_action\_steps & 6 & 6 \\ 
num\_inference\_steps & 100 & 10 \\ 
dataloader.batch\_size & 128 & 256 \\ 
dataloader.num\_workers & 0 & 8 \\ 
dataloader.shuffle & True & True \\ 
dataloader.pin\_memory & True & True \\ 
dataloader.persistent\_workers & False & False \\ 
optimizer.\_target\_ & torch.optim.AdamW & torch.optim.AdamW \\ 
optimizer.lr & 1.0e-4 & 1.0e-4 \\ 
optimizer.betas & [0.95, 0.999] & [0.95, 0.999] \\ 
optimizer.eps & 1.0e-8 & 1.0e-8 \\ 
optimizer.weight\_decay & 1.0e-6 & 1.0e-6 \\ 
training.lr\_scheduler & cosine & cosine \\ 
training.lr\_warmup\_steps & 500 & 500 \\ 
training.num\_epochs & 300 & 3000 \\ 
training.gradient\_accumulate\_every & 1 & 1 \\ 
training.use\_ema & True & True \\ 
\bottomrule
\end{tabular}
\vspace{-3pt}
\caption{Baselines Settings. Hyper-parameter Settings for Training and Deployment of DP and DP3 Algorithms.}
\label{tab:baseline_setup}
\end{table*}

\end{document}